\documentclass[conference]{IEEEtran}
\IEEEoverridecommandlockouts

\usepackage{cite}
\usepackage{amsmath,amssymb,amsfonts}
\usepackage{algorithmic}
\usepackage{graphicx}
\usepackage{textcomp}
\usepackage{xcolor}
\usepackage{algorithm}
\usepackage{algorithmic}
\usepackage{multirow}
\usepackage[hidelinks]{hyperref}
\usepackage{pifont}
\def\BibTeX{{\rm B\kern-.05em{\sc i\kern-.025em b}\kern-.08em
    T\kern-.1667em\lower.7ex\hbox{E}\kern-.125emX}}

\begin{document}
\title{Efficient Federated Learning with Encrypted Data Sharing for Data-Heterogeneous Edge Devices}

\author{\IEEEauthorblockN{Hangyu Li$^{1}$,
Hongyue Wu$^{1,2,3}$, 
Guodong Fan$^{1}$, 
Zhen Zhang$^{1}$,
Shizhan Chen$^{1}$\IEEEauthorrefmark{1}\thanks{*Corresponding author. },
Zhiyong Feng$^{1}$\thanks{Code available at: https://github.com/XiaoziLee/FedEDS.}
}
\IEEEauthorblockA{$^1$\textit{College of Intelligence and Computing, Tianjin University, Tianjin, China}}
\IEEEauthorblockA{$^2$\textit{State Key Lab. for Novel Software Technology, Nanjing University, Nanjing, China}}
\IEEEauthorblockA{$^3$\textit{Yunnan Key Lab. of Service Computing, Yunnan University of Finance and Economics, Kunming, China}}
\{hangyulee, hongyue.wu, guodongfan, zhang\_z, shizhan, zyfeng\}@tju.edu.cn
}

\maketitle

\begin{abstract}
As privacy protection gains increasing importance, more models are being trained on edge devices and subsequently merged into the central server through Federated Learning (FL). However, current research overlooks the impact of network topology, physical distance, and data heterogeneity on edge devices, leading to issues such as increased latency and degraded model performance.
To address these issues, we propose a new federated learning scheme on edge devices that called Federated Learning with Encrypted Data Sharing(FedEDS).
FedEDS uses the client model and the model's stochastic layer to train the data encryptor. The data encryptor generates encrypted data and shares it with other clients. The client uses the corresponding client's stochastic layer and encrypted data to train and adjust the local model. FedEDS uses the client's local private data and encrypted shared data from other clients to train the model.
This approach accelerates the convergence speed of federated learning training and mitigates the negative impact of data heterogeneity, making it suitable for application services deployed on edge devices requiring rapid convergence. 
Experiments results show the efficacy of FedEDS in promoting model performance. 
\end{abstract}


\section{Introduction}
Federated Learning is a distributed machine learning framework designed to build a globally shared model through collaborative training across devices or nodes while preserving data privacy \cite{mcmahan2017communication,yang2019federated,yang2020federated}. Unlike traditional centralized training methods, Federated Learning eliminates the need to upload local data to a central server, instead decentralizing the training process to participating nodes. Global optimization is achieved through the exchange of model parameters or gradients. This approach is widely applied in data-sensitive scenarios, such as mobile devices \cite{yi2024fedpe,jiang2022fedmp,he2020group}, healthcare, and financial fields \cite{yang2019federated}, striking a balance between privacy preservation and computational efficiency.

One key challenge in Federated Learning is the heterogeneity of data across different nodes. This phenomenon, known as data heterogeneity\cite{Li_2020}, refers to the fact that each device or edge node may have access to a different distribution of data, which can significantly vary in terms of quality and type. For example, in healthcare, different hospitals may have distinct patient demographics and varying levels of data collection practices, which can lead to biased or unrepresentative local models. This heterogeneity can hinder the convergence of the global model and lead to inefficiencies in training. To address this, various techniques have been developed to account for the diversity in local datasets, such as adjusting for local model updates or incorporating federated averaging strategies. Nevertheless, addressing data heterogeneity remains a significant obstacle in achieving optimal performance and generalization in Federated Learning systems.
\begin{figure}[t]
\centering
\includegraphics[width=0.9\columnwidth]{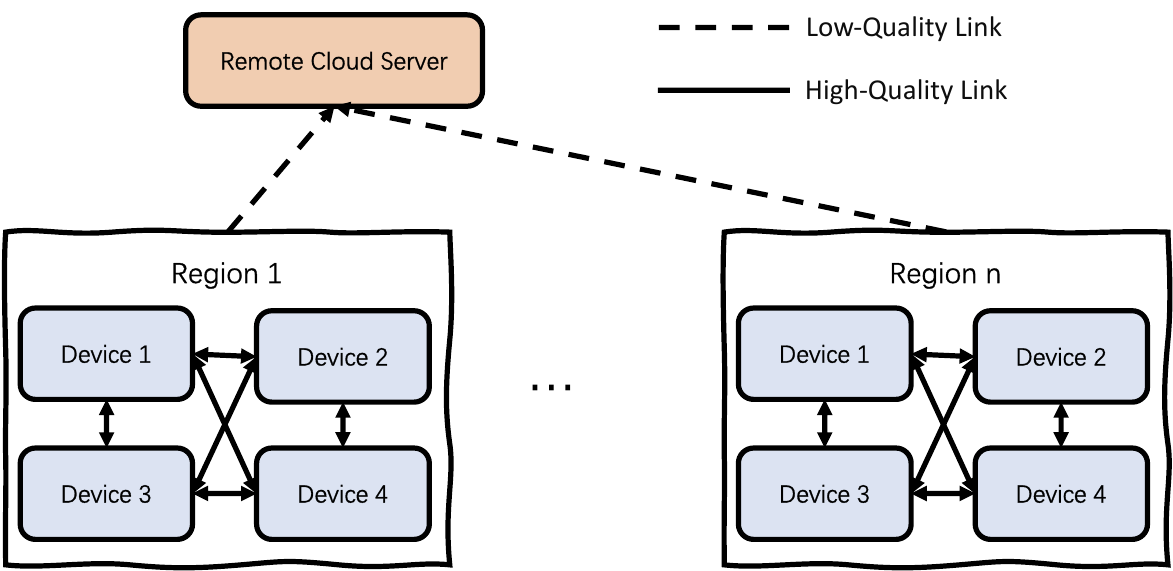}
\caption{Communication in Edge Computing. Communication between cloud servers and edge devices is often costly, slow, and unstable, due to physical distance and other environmental constraints. In contrast, communication between edge devices within the same geographic region is typically more convenient. Our proposed method introduces an information exchange mechanism among edge devices within the same region to share encrypted data, effectively enhancing the performance of the federated learning model.}
\label{fig1}
\end{figure}

In practical applications, combining edge computing with Federated Learning can significantly improve efficiency and privacy. As shown in Fig.\ref{fig1}, edge server nodes are physically close and can communicate directly through high-quality links. However, communication between edge servers and the central server often suffers from poor conditions, including high latency and instability. In traditional Federated Learning, model parameters or gradients need to be frequently transmitted to the central server for global aggregation. High communication latency not only prolongs training time but also forces edge servers to wait for aggregation results, significantly reducing the overall efficiency of Federated Learning in edge computing scenarios \cite{lim2020federated}. Additionally, the data distribution on edge servers is often highly heterogeneous \cite{kairouz2021advances,li2021survey}, further complicating model training and hindering the convergence of the global model. Existing studies\cite{zhao2018federated} suggest that incorporating a limited amount of shared data can significantly improve model performance in Federated Learning. However, sharing data may also introduce privacy risks, as it could potentially reveal sensitive information about other clients' private datasets.

To address above challenges, we propose a novel method, \textbf{Fed}erated Learning with \textbf{E}ncrypted \textbf{D}ata \textbf{S}haring, \textbf{FedEDS}. FedEDS leverages high-speed direct links between edge servers to exchange encrypted data information for knowledge transfer among edge server clients without compromising data privacy. These encrypted data exchanges effectively mitigate the adverse effects of data heterogeneity on the Federated Learning training process, resulting in more efficient model performance on edge servers. Additionally, FedEDS employs a piecewise epoch annealing strategy to further reduce communication rounds with the central server. FedEDS significantly reduces the impact of central server communication delays on training efficiency while enhancing the overall performance of Federated Learning in edge computing scenarios.
Our main contributions are as follows:
\begin{itemize}
    \item To address the issue of high latency in the central server within edge computing scenarios, we propose a design concept that takes advantage of direct communication between edge servers to enhance the efficiency of federated learning.

    \item We propose FedEDS, which combines encrypted data sharing between edge servers with a piecewise epoch annealing strategy to effectively mitigate the negative impact of data heterogeneity on model training while accelerating convergence.
    
    \item Through experimental validation, FedEDS consistently demonstrates outstanding performance in terms of model convergence speed and performance improvement, showcasing its advantages in edge computing scenarios comprehensively.
\end{itemize}

\section{Related Work}
\subsection{Data Heterogeneity in Federated Learning}
In federated learning, data heterogeneity is a critical challenge that affects model performance. Existing research primarily addresses this issue from two perspectives: (1) reducing the impact of heterogeneity on the optimization process by correcting the bias between local and global models, and (2) improving the model aggregation mechanism to accommodate the differences in client updates.

The first approach aims to constrain the optimization paths of local models, reducing the bias between local and global models. MOON \cite{li2021model} introduces a contrastive learning-based mechanism that enhances the similarity between local and global model representations while minimizing the variations of local models across training rounds. This ensures that the local optimization direction aligns more closely with the global objective. FedProx\cite{li2020federated} adds a regularization term to the local training objective function to limit the deviation of local models from the global model, effectively mitigating the negative impact of non-independent and identically distributed (non-IID) data on model convergence.

The second approach focuses on improving the global model aggregation mechanism to better handle the differences in client updates. FedNova\cite{wang2020tackling} normalizes local updates from clients, addressing the impact of uneven optimization steps and data distributions across clients on global model updates, thereby achieving a fairer aggregation process. FedAvgM \cite{hsu2019measuringeffectsnonidenticaldata} introduces a momentum mechanism into the classic FedAvg\cite{mcmahan2017communication} algorithm, utilizing historical update information to smooth the global model's update direction. This reduces oscillations caused by differences in client updates, thereby improving the convergence efficiency and robustness of the global model.

\subsection{Data Generation in Federated Learning}
Previous works \cite{zhang2022dense, zhu2021data, hao2021towards, zhang2022fine} have explored generating data from models uploaded by clients to the server to fine-tune the global model. However, this approach faces challenges, especially in federated learning scenarios where secure aggregation techniques \cite{so2022lightsecagg} are often employed. These techniques ensure that the server only receives the aggregated global model and cannot access individual client updates.

\subsection{Communication-Efficient Federated Learning in Edge Devices}

The efficiency of federated learning on edge devices is constrained by the communication capabilities of both the edge devices and the central server responsible for aggregating the model. Many related studies\cite{jiang2022modelpruningenablesefficient, jia2024dapperfldomainadaptivefederated} have proposed using model pruning methods to reduce the communication load between edge devices and the central server, while others\cite{9660377, zhu2024efficientmodelcompressionhierarchical, xia2024bayesianfederatedmodelcompression} have suggested using model compression techniques to reduce the model size, thereby improving the efficiency of federated learning.

\section{Preliminary}
Federated Learning is a decentralized machine learning paradigm where multiple clients collaboratively train a global model under the coordination of a central server while maintaining the privacy of local data. This section introduces the basic definitions and mathematical formulation of FL.

Assume there are \( K \) clients, each holding a local dataset \( \mathcal{D}_k \) of size \( n_k \), where \( k \in \{1, 2, \ldots, K\} \). The objective is to learn a global model \( \theta_G \) by minimizing the following overall objective function:

\begin{equation}
\min_{\theta_G} \mathcal{L}_G(\theta_G) =\sum_{k=1}^K  p_k \mathcal{L}_k(\theta_k),\label{eq1}
\end{equation}
where \(p_k = \frac{n_k}{n}\), \( n = \sum_{k=1}^K n_k \) is the total number of data samples, and \( \mathcal{L}_k(\theta_k) = \frac{1}{n_k} \sum_{(x,y) \in \mathcal{D}_k}^{n_k} \mathcal{L}(\theta_k; x, y) \) represents the local loss function on client \( k \). Here, \( \mathcal{L}(\theta_k; x, y) \) denotes the loss for a single data point \( (x, y) \).

This formulation allows each client to contribute to the optimization process proportionally to the size of its local dataset while ensuring that the training respects the privacy of individual data points.

\section{Methodology}

\begin{figure*}[t]
\centering
\includegraphics[width=0.9\textwidth]{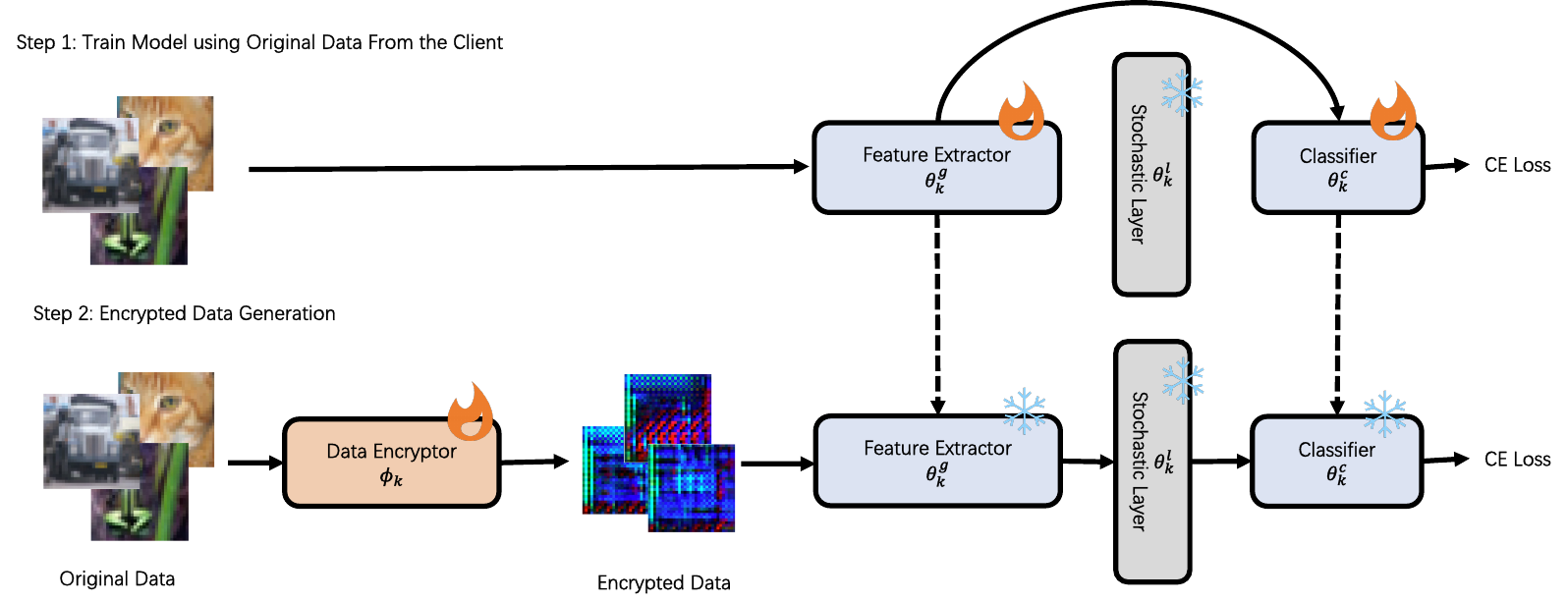} 
\caption{In Step 1 of the figure, the client model is trained normally using its local dataset, during which the forward propagation of the model skips the stochastic layer. Step 2 involves the training process of the client’s data encryptor. In this step, the local data of the client is used as the input to the data encryptor, and the output of the encryptor serves as the input to the pre-trained client model. During this process, the client model utilizes the stochastic layer for forward propagation, with the parameters of this layer remaining frozen throughout. The role of the stochastic layer is to introduce noise perturbations to the features extracted by the feature extractor from the original images. The data encryptor aims to output encrypted data that can counteract such perturbations. This ensures that the features output by the feature extractor and the stochastic layer when processing the encrypted data remain consistent with those derived from the original images processed solely by the feature extractor, thereby enabling the classifier to make accurate predictions.}
\label{fig2}
\end{figure*}

In our proposed method, each client \( k \) possesses model parameters \( \theta_k \), which can be further divided into three components: trainable feature extractor parameters \( \theta_k^g \), trainable classifier parameters \( \theta_k^c \), and frozen parameters \( \theta_k^l \) of the stochastic layer, which do not participate in training.

\subsection{Encrypted Data Generation}
Before federated learning begins, the participating edge devices first generate encrypted data. As shown in Step 1 of Fig. \ref{fig2}, during the encrypted data feature generation process, client \( k \) optimizes the model parameters \( \theta_k^g \) and \( \theta_k^c \) by minimizing the loss function \( \mathcal{L}_{c} \) on its local dataset. The optimization objective can be expressed as:

\begin{equation}
\min_{\theta_k^g, \theta_k^c} \mathcal{L}_{c}(\theta_k^g, \theta_k^c; \mathcal{D}_k) = \min_{\theta_k^g, \theta_k^c} \sum_{(x, y) \in \mathcal{D}_k}^{n_k} \mathcal{L}_{CE}(y, f(x; \theta_k^g, \theta_k^c)),\label{eq2}
\end{equation}
where \( \mathcal{D}_k \) represents the local dataset of the \( k \)-th client, \( n_k \) is the number of samples in the local dataset of client \( k \), \(y\) is label, and \( \mathcal{L}_{CE} \) denotes the cross-entropy loss.

It is important to note that the parameters of the stochastic layer, \( \theta_k^l \), do not participate in the forward propagation or parameter updates during the training process described above.

\begin{figure*}[t]
\centering
\includegraphics[width=1.0\textwidth]{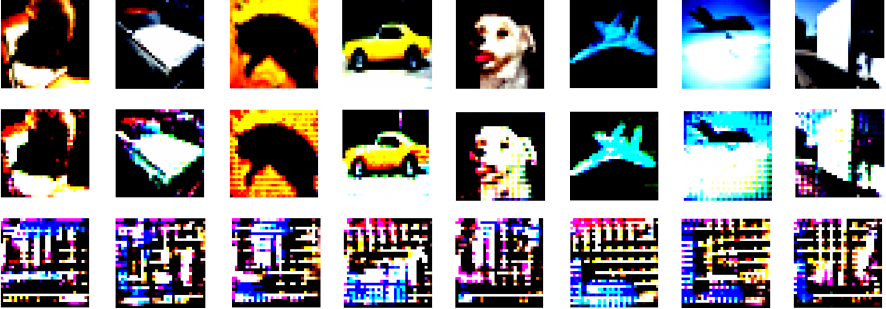} 
\caption{The first row of the figure showcases original images from the CIFAR-10\cite{krizhevsky2009learning} dataset, while the second and the third row visualizes the impact of the stochastic layer on privacy protection during the encrypted data generation process. The second row demonstrates the output of the data encryptor without incorporating the client model's stochastic layer, showing that the result is nearly indistinguishable from the original images. In contrast, when the data encryptor integrates the stochastic layer of the client model during training, the generated encrypted data becomes unrecognizable to the human eye. Although these encrypted data are not directly interpretable by humans, they can provide more useful information for other client models in federated learning, thereby enhancing learning performance and mitigating the adverse effects of data heterogeneity.}
\label{fig3}
\end{figure*}

In our method, each client \( k \) has a data encryptor that leverages the parameters \( \theta_k^l \) of the stochastic layer in the model to generate encrypted data. These encrypted data are indistinguishable to the human eye but can effectively convey knowledge from the model. As illustrated in Step 2 of Fig.\ref{fig2}, the training objective of the data encryptor can be expressed as:

\begin{equation}
\min_{\phi_k} \mathcal{L}_{c}(\phi_k; \mathcal{D}_k) = \min_{\phi_k} \sum_{(x, y) \in \mathcal{D}_k}^{n_k} \mathcal{L}_{CE}(y, f(g(x; \phi_k); \theta_k^g, \theta_k^c, \theta_k^l)), \label{eq3}
\end{equation}
where \( \phi_k \) denotes the parameters of the data encryptor for client \( k \).

During the training of the data encryptor, the parameters \( \theta_k^l \) of the stochastic layer in the client model are involved in forward propagation, while all the model parameters \( \theta_k = \{ \theta_k^g, \theta_k^c, \theta_k^l \} \) remain frozen.

Once the data encryptor has been trained, it is used to generate encrypted data from the local dataset \( \mathcal{D}_k \) of client \( k \). These encrypted data, along with the parameters \( \theta_k^l \) of the client’s stochastic layer, are shared with other clients to transfer the knowledge encapsulated in the model of client \( k \).

Each client \( k \) constructs and maintains an encrypted dataset \( \mathcal{D}_k^F \) derived from its local dataset:

\begin{equation}
\mathcal{D}_k^F = \{ (g(x; \phi_k), f(g(x; \phi_k); \theta_k^g, \theta_k^c, \theta_k^l)) \mid (x, y) \in \mathcal{D}_k \},\label{eq4}
\end{equation}

Throughout the training of the data encryptor and the generation of the encrypted dataset, the parameters of the stochastic layer are frozen while actively participating in forward propagation. Algorithm \ref{algorithm1} summarizes the procedure of encrypted data generation. Fig.\ref{fig3} shows the generated encrypted data.

\begin{algorithm}[tb]
\caption{Federated Encrypted Data Generation Based on Stochastic Layer}
\label{algorithm1}
\textbf{Input}: Client ID \( k \), global model from server \( \theta_G^{g,t},\theta_G^{c,t} \) in communication round \( t \), stochastic layer parameters of the client \( \theta_k^{l}\), local epochs \(E_c, E_g\)\\
\textbf{Output}: Encrypted datasets \(\mathcal{D}_k^{F}\)\\
\textbf{FedEncryptedDataGenerate(\(k, \theta_G^{g,t}, \theta_G^{c,t}, \theta_k^{l}\))}:
\begin{algorithmic}[1]
\STATE \(\theta_G^{g,t}, \theta_G^{c,t}\) initialize local model \(\theta_k^{g}, \theta_k^{c} \)
\FOR{each local epoch \(e = 1,2,...,E_c\)}
    \STATE \( \theta_k^{g}, \theta_k^{c} \leftarrow \) SGD update use Eq\eqref{eq2}
\ENDFOR
\FOR{each local epoch \(e= 1,2,...,E_g\)}
    \STATE \( \phi_k\leftarrow \) SGD update use Eq\eqref{eq3}
\ENDFOR
\STATE Generate Encrypted datasets \(\mathcal{D}_k^{F}\)  use Eq\eqref{eq4}
\STATE Restore local model \(\theta_k^{g}, \theta_k^{c} \) to \(\theta_G^{g,t}, \theta_G^{c,t}\)
\STATE \textbf{return} \(\mathcal{D}_k^{F}\)
\end{algorithmic}
\end{algorithm}

\subsection{Piecewise Epoch Annealing Strategy}
In the early stages of federated learning, model parameters are still in the exploration phase. During this period, using a larger number of local epochs allows each client to perform more iterations of local training, quickly optimizing the initial model parameters and accelerating the convergence of the overall federated learning system. As training progresses, the model becomes more stable, and the magnitude of parameter updates decreases. At this stage, reducing the number of epochs can help prevent overfitting and improve training efficiency. To achieve dynamic adjustment of the local training iterations (i.e., the number of epochs), we propose a strategy where a large number of epochs \( E_{\text{max}} \) is used during the first \( T_{\alpha}^{\text{turn}} \) communication rounds. Afterward, the number of epochs gradually decreases, eventually being fixed to \(E_{\text{min}}\) in later rounds. The number of local epochs \( E_t \) in communication round \( t \) is defined as follows:
\begin{equation}
E_t = 
\begin{cases}
E_{\text{max}}, & t \leq T_{\alpha}^{\text{turn}}, \\
E_{\text{max}} - \left\lfloor \frac{(E_{\text{max}} - E_{\text{min}}) \cdot (t - T_{\alpha}^{\text{turn}})}{(T_{\beta}^{\text{turn}} - T_{\alpha}^{\text{turn}})} \right\rfloor, & T_{\alpha}^{\text{turn}} < t \leq T_{\beta}^{\text{turn}}, \\
E_{\text{min}}, & t > T_{\beta}^{\text{turn}},
\end{cases}\label{eq5}
\end{equation}
where \( T_{\alpha}^{\text{turn}} \) and \( T_{\beta}^{\text{turn}} \) are predefined turning points that determine the number of communication rounds required for the epochs to decrease from \( E_{\text{max}} \) to \(E_{\text{min}}\).

This piecewise epoch annealing strategy ensures an optimized training process that balances accuracy and resource utilization. It adapts to the varying needs of the model at different stages of training, enhancing both convergence speed and overall efficiency.

\subsection{Knowledge Transfer by Encrypted Data Sharing}
During the knowledge transfer phase, client \( k \) trains its model using both its local dataset and the encrypted datasets \( \{\mathcal{D}_i^F\}_{i=1, i \neq k}^K \) from other clients. This process enhances client \( k \)’s ability to recognize patterns in private data samples from other clients. When training the model with an encrypted dataset \( \mathcal{D}_i^F \) from client \( i \), the stochastic layer's parameters \( \theta_k^l \) of client \( k \)’s model are replaced with \( \theta_i^l \). The combined loss function, incorporating both local and encrypted datasets, is then used to optimize the client’s model.

For the encrypted dataset \( \mathcal{D}_i^F \) sent by client \( i \), the loss function is defined as:

\begin{equation}
\mathcal{L}_{dis}(\theta_k^g; \theta_k^c; \theta_i^l; \mathcal{D}_i^F) = \sum_{(x, y) \in \mathcal{D}_i^F}^{\left| \mathcal{D}_i^F \right|} \mathcal{L}_{KL}(y, f(x; \theta_k^g; \theta_k^c; \theta_i^l)),\label{eq6}
\end{equation}
where \( \mathcal{L}_{KL} \) is the Kullback-Leibler (KL) divergence loss, \( \left| \mathcal{D}_i^F \right| \) is the size of the encrypted dataset \( \mathcal{D}_i^F \). 

Combining the local dataset \( \mathcal{D}_k \) and the encrypted datasets \( \{\mathcal{D}_i^F\}_{i=1, i \neq k}^K \), the total loss function for client \( k \) is:

\begin{equation}
\begin{aligned}
& \min_{\theta_k^g, \theta_k^c} \mathcal{L}_k(\theta_k^g; \theta_k^c; \{\theta_i^l\}_{i=1, i \neq k}^K; \mathcal{D}_k; \{\mathcal{D}_i^F\}_{i=1, i \neq k}^K) \\ 
& = \min_{\theta_k^g, \theta_k^c} \left[ \lambda_c \underbrace{\mathcal{L}_c(\theta_k^g, \theta_k^c; \mathcal{D}_k)}_{\text{Local knowledge}} + \lambda_{dis} \underbrace{\sum_{i \neq k}^{K - 1} \mathcal{L}_{dis}(\theta_k^g; \theta_k^c; \theta_i^l; \mathcal{D}_i^F))}_{\text{Knowledge from all other clients}} \right],
\end{aligned}\label{eq7}
\end{equation}
where \( \lambda_c \) and \( \lambda_{dis} \) are scaling factors for the losses.

In practice, the encrypted datasets \( \{\mathcal{D}_i^F\}_{i=1, i \neq k}^K \) often contain significantly more samples than the local dataset \( \mathcal{D}_k \), especially when the number of clients \( K \) is large. This imbalance can lead to models focusing more on encrypted datasets, neglecting local data, and increasing computational burden. To address this, client \( k \) performs uniform sampling of client indices to select a subset of encrypted datasets for training. The revised optimization objective is:

\begin{equation}
\begin{aligned}
& \min_{\theta_k^g, \theta_k^c} \mathcal{L}_k(\theta_k^g; \theta_k^c; \{\theta_i^l\}_{i=1, i \neq k}^K; \mathcal{D}_k; \{\mathcal{D}_i^F\}_{i=1, i \neq k}^K)  \\ 
& = \min_{\theta_k^g, \theta_k^c} \left[ \lambda_c \underbrace{ \mathcal{L}_c(\theta_k^g, \theta_k^c; \mathcal{D}_k)}_{\text{Local knowledge}} + \lambda_{dis} \underbrace{\mathcal{L}_{dis}(\theta_k^g; \theta_k^c; \theta_{i^*}^l; \mathcal{D}_{i^*}^F)}_{\text{Knowledge from one other client}} \right], \\
& \text{subject to } i^* \sim \mathcal{U}(\{1, \dots, K\} \setminus \{k\}),
\end{aligned}\label{eq8}
\end{equation}
where \( i^* \) is a randomly sampled index from a uniform distribution excluding \( k \). 

\begin{figure*}[t]
\centering
\includegraphics[width=0.9\textwidth]{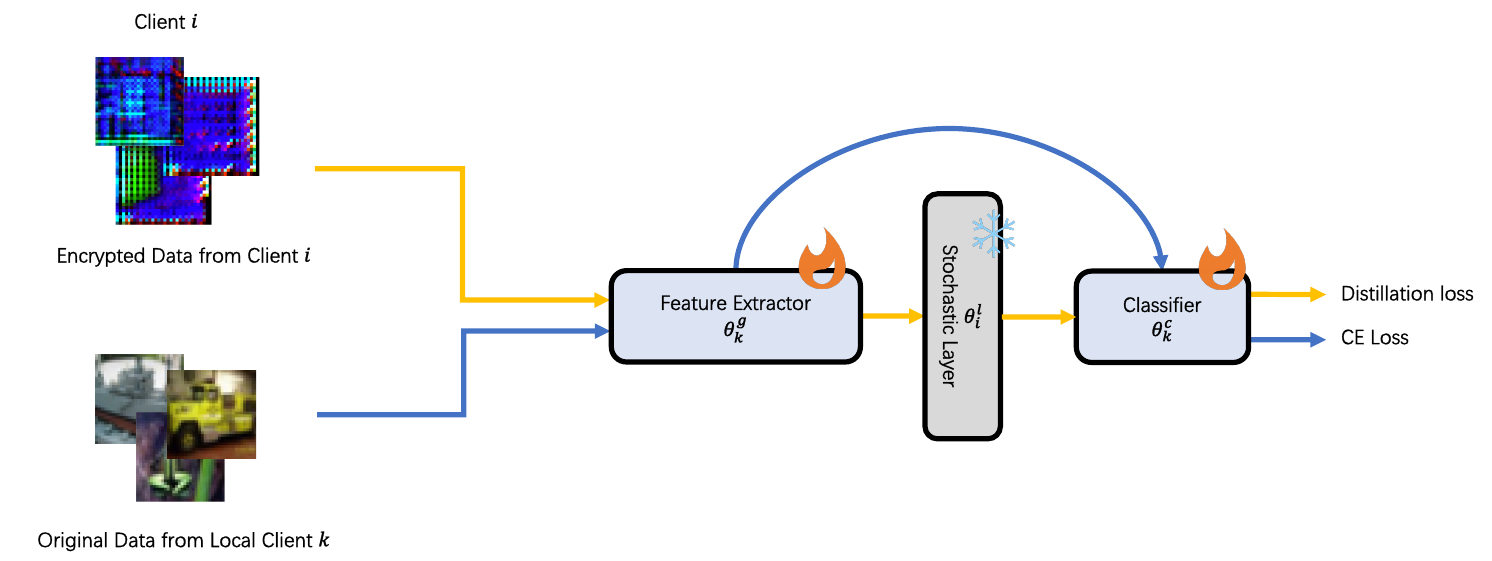} 
\caption{The figure illustrates the process of optimizing the model on client \( k \) using both its local private data and encrypted data from other clients. The yellow lines represent the forward propagation of encrypted data through the client model, while the blue lines indicate the forward propagation of the original local data through the client model.}
\label{fig4}
\end{figure*}

Fig.\ref{fig4} provides a detailed illustration of the process of optimizing the model using encrypted data from other clients combined with local raw data.

To prevent overfitting to encrypted data and encourage the model to prioritize local data, the scaling factors \( \lambda_c \) and \( \lambda_{dis} \) are dynamically adjusted. Let \( t \) denote the current communication round, then:

\begin{equation}
\lambda_c =
\begin{cases}
1, & \frac{1}{1 + e^{-m(t - 1)}} \geq 1 - \epsilon, \\
\frac{1}{1 + e^{-m(t - 1)}}, & \text{otherwise},
\end{cases}\label{eq9}
\end{equation}

\begin{equation}
\lambda_{dis} =
\begin{cases}
0, & \frac{e^{-m(t - 1)}}{1 + e^{-m(t - 1)}} < \epsilon, \\
\frac{e^{-m(t - 1)}}{1 + e^{-m(t - 1)}}, & \text{otherwise},
\end{cases}\label{eq10}
\end{equation}
where \( \lambda_c + \lambda_{dis} = 1 \), \( \forall t \in [1, T] \). Initially, \( \lambda_c = \lambda_{dis} = 0.5 \), and as \( t \to T \), \( \lambda_c \to 1 \) and \( \lambda_{dis} \to 0 \). This ensures \( \lambda_{dis} \) decays rapidly in early stages, while \( \lambda_c \) increases quickly. When \( \lambda_{dis} < \epsilon \), it is set to 0, and \( \lambda_c \) is set to 1. The parameter \( m \) controls the speed of change for \( \lambda_c \) and \( \lambda_{dis} \).

Algorithm \ref{algorithm2} summarizes the procedure of knowledge transfer by encrypted data
sharing and piecewise epoch annealing strategy.

\begin{algorithm}[tb]
\caption{Federated Knowledge Transfer using Encrypted Data Datasets}
\label{algorithm2}
\textbf{Input}: Client ID \( k \), global model from server \( \theta_G^{g,t},\theta_G^{c,t} \) in communication round \( t \), stochastic layer parameters of all clients \(\{\theta_k^{l}\}_{k=1}^{K}\), Encrypted datasets from other clients \(\{\mathcal{D}_i^{F}\}_{i=1,i \neq k}^{K} \), \( T_{\alpha}^{turn}, T_{\beta}^{turn}, E_{max}\)\\
\textbf{Output}: \(\theta_k^{g,t+1}, \theta_k^{c,t+1}\)\\
\textbf{FedKnowTrans(\(k, \theta_G^{g,t}, \theta_G^{c,t},\{\theta_k^{l}\}_{k=1}^{K},\{\mathcal{D}_i^{F}\}_{i=1,i \neq k}^{K}\))}
\begin{algorithmic}[1]
\STATE \(\theta_G^{g,t}, \theta_G^{c,t}\) initialize local model \(\theta_k^{g,t}, \theta_k^{c,t} \)
\STATE Compute \( E_t \) using Eq\eqref{eq5} with \( T_{\alpha}^{turn}, T_{\beta}^{turn}, E_{max} \)
\STATE Compute \( \lambda_{c}, \lambda_{dis} \) using Eq\eqref{eq9} and Eq\eqref{eq10} with \( t \)
\FOR{each local epoch \(e = 1,2,...,E_t\)}
    \STATE \( \theta_k^{g,t+1}, \theta_k^{c,t+1} \leftarrow \) SGD update use Eq\eqref{eq8}/FedProx use Eq\eqref{eq11} with \( \lambda_{c}, \lambda_{dis} \)
\ENDFOR
\STATE \textbf{return} \(\theta_k^{g,t+1}, \theta_k^{c,t+1}\)
\end{algorithmic}
\end{algorithm}

\begin{algorithm*}[tb]
\caption{Federated Learning with Encrypted Data Sharing(FedEDS)}
\label{algorithm3}
\textbf{Input}: Number of clients \(K\), total number of communication rounds \(T\), local epoch \(E_c, E_g\)\\
\textbf{Server Executes}:
\begin{algorithmic}[1] 
\STATE Server initialize the global model \(\theta_G^{g,0},\theta_G^{c,0}\) and the model stochastic layer parameters of each client \( \{\theta_k^{l}\}_{k=1}^{K}\)
\STATE Server broadcast \(\theta_G^{g,0},\theta_G^{c,0}\) to all clients, distribute \( \{\theta_k^{l}\}_{k=1}^{K}\) to corresponding clients
\STATE Each client initialize feature generator parameters \(\{\phi_k\}_{k=1}^{K}\)
\FOR{each client \(k = 1,2,...,K\) \textbf{in parallel}}
    \STATE \(\mathcal{D}_k^{F} \leftarrow \) \textbf{FedEncryptedDataGenerate(\(k, \theta_G^{g,0},\theta_G^{c,0},\theta_k^{l}\))}
    \STATE Broadcast \( \mathcal{D}_k^{F} \) to all other participating clients
\ENDFOR
\FOR{each round \(t = 1,2,...,T \)}
    \FOR{each client \(k = 1,2,...,K\) \textbf{in parallel}}
        \STATE \(\theta_k^{g,t+1},\theta_k^{c,t+1} \leftarrow \) \textbf{FedKnowTrans(\(k, \theta_G^{g,t},\theta_G^{c,t},\{\theta_k^{l}\}_{k=1}^{K},\{\mathcal{D}_i^{F}\}_{i=1,i \neq k}^{K}\))}
    \ENDFOR
    \STATE \( \theta_G^{g, t+1}, \theta_G^{c, t+1} = Aggregate(\{\theta_k^{g,t+1}, \theta_k^{c,t+1}\}_{k=1}^{K})\)
    \STATE Broadcast \( \{\theta_G^{g, t+1}, \theta_G^{c, t+1} \} \) to all participating clients
\ENDFOR
\end{algorithmic}
\end{algorithm*}

\section{Discussion}
\subsection{FedEDS Combined With Other Methods}
The following section explains the pseudocode for combining FedEDS with existing federated learning algorithms. Algorithms \ref{algorithm3} illustrate the process of integrating FedEDS with FedAvg\cite{mcmahan2017communication} and FedProx\cite{li2020federated}. The process of combining FedProx\cite{li2020federated} with FedEDS is similar to that of combining FedAvg\cite{mcmahan2017communication} with FedEDS, with the main difference being that FedProx\cite{li2020federated} includes an additional proximal term in the local loss function compared to FedAvg\cite{mcmahan2017communication}.
\begin{equation}
\begin{aligned}
\min_{\theta_k^g, \theta_k^c} \lambda_{c} \mathcal{L}_{c}(\theta_k^g, \theta_k^c;\mathcal{D}_k) + \lambda_{dis} \sum_{i \neq k}^{K - 1} \mathcal{L}_{dis}(\theta_k^{g};\theta_k^{c};\theta_i^{l};\mathcal{D}_{i}^{F}) \\ + \frac{\mu}{2} \left( \|\theta_k^g - \theta_G^g\|_2^2 + \|\theta_k^c - \theta_G^c\|_2^2 \right) ,
\end{aligned}
\label{eq11}
\end{equation}
the hyperparameter \(\mu\) serves to adjust the regularization strength in local optimization on the client side, aiming to address the challenges posed by data distribution heterogeneity across different clients.

For the FedNova\cite{wang2020tackling}algorithm, it is necessary to calculate \(\|a_k\|_1\) for each client \(k\). In our experiments, we use the momentum SGD optimizer.

\begin{algorithm}[tb]
\caption{Federated Learning using Encrypted Data with FedNova}
\label{algorithm4}
\textbf{Input}: Number of clients \(K\), total number of communication rounds \(T\), local epoch \(E_c, E_g\), learning rate \(\eta\), momentum factor \( \rho\)\\
\textbf{Server Executes}:
\begin{algorithmic}[1] 
\STATE Server initialize the global model \(\theta_G^{g,0},\theta_G^{c,0}\) and the model stochastic layer parameters of each client \( \{\theta_k^{l}\}_{k=1}^{K}\)
\STATE Server broadcast \(\theta_G^{g,0},\theta_G^{c,0}\) to all clients, distribute \( \{\theta_k^{l}\}_{k=1}^{K}\) to corresponding clients
\STATE Each client initialize feature generator parameters \(\{\phi_k\}_{k=1}^{K}\)
\FOR{each client \(k = 1,2,...,K\) \textbf{in parallel}}
    \STATE \(\mathcal{D}_k^{F} \leftarrow \) \textbf{FedFeatGenerate(\(k, \theta_G^{g,0},\theta_G^{c,0},\theta_k^{l}\))}
    \STATE Broadcast \( \mathcal{D}_k^{F} \) to all other participating clients
\ENDFOR
\FOR{each round \(t = 1,2,...,T \)}
    \FOR{each client \(k = 1,2,...,K\) \textbf{in parallel}}
        \STATE \(\theta_G^{g,t}, \theta_G^{c,t}\) initialize local model \(\theta_k^{g,t}, \theta_k^{c,t} \)
        \STATE Compute \( E_t \) using Eq\eqref{eq5}
        \STATE Compute \( \lambda_{c}, \lambda_{dis} \) using Eq\eqref{eq9} and Eq\eqref{eq10}
        \FOR{each local epoch \(e = 1,2,...,E_t\)}
            \STATE \( \theta_k^{g,t+1}, \theta_k^{c,t+1} \leftarrow \) SGD update use Eq\eqref{eq8}
        \ENDFOR
        \STATE \( \|a_k\|_1 \leftarrow \) Compute using Eq\eqref{eq12} with \( \eta \) and \(\rho\)
        \STATE \(\Delta\theta_k^{g,t+1}\leftarrow \frac{\theta_k^{g,t+1} - \theta_k^{g,t}}{\eta\|a_k\|_1} \)
        \STATE \(\Delta\theta_k^{c,t+1}\leftarrow \frac{\theta_k^{c,t+1} - \theta_k^{c,t}}{\eta\|a_k\|_1} \)
        \STATE Send \(\Delta\theta_k^{g,t+1}, \Delta\theta_k^{c,t+1}, \|a_k\|_1\) to server
    \ENDFOR
    \STATE \(\theta_G^{g,t+1} \leftarrow \theta_G^{g,t} - \frac{\sum_{k}^{K}\|a_k\|_1}{K} \sum_{k}^{K}\frac{\Delta\theta_k^{g,t+1}}{K}  \)
    \STATE \(\theta_G^{c,t+1} \leftarrow \theta_G^{c,t} - \frac{\sum_{k}^{K}\|a_k\|_1}{K} \sum_{k}^{K}\frac{\Delta\theta_k^{c,t+1}}{K}  \)
    \STATE Broadcast \( \{\theta_G^{g, t+1}, \theta_G^{c, t+1} \} \) to all participating clients
\ENDFOR
\end{algorithmic}
\end{algorithm}

\begin{equation}
\|a_k\|_1 = \frac{1}{1-\rho} \left[ E_t - \frac{\rho (1 - \rho^{E_t})}{1 - \rho} \right],\label{eq12}
\end{equation}
here, \(a_k\) is a vector of length \(E_t\), where \(E_t\) represents the number of local updates performed by the client during the \(t\)-th communication round. \(a_k\) is the weight vector of accumulated gradients during the local updates on the client, and \(\|a_k\|_1\) is the sum of these weights. This sum is used to normalize the local update gradients, aiming to eliminate objective inconsistency. Algorithm \ref{algorithm4} illustrates the process of combining FedNova\cite{wang2020tackling}with FedEDS.

FedEDS exhibits exceptional scalability. Beyond its compatibility with FedAvg\cite{mcmahan2017communication}, FedProx\cite{li2020federated}, and FedNova\cite{wang2020tackling}, it can seamlessly integrate with a wide range of other federated learning techniques.

\subsection{Theoretical Convergence Analysis of FedEDS}
We present the theoretical convergence analysis of our method. There are four conventional assumptions in Federated Learning theoretical literature\cite{wang2020tackling}. To make writing easier, we use \(\mathcal{L}_k(\theta)\) to denote \(\mathcal{L}_k(\theta_k^g; \theta_k^c; \{\theta_i^l\}_{i=1, i \neq k}^K; \mathcal{D}_k; \{\mathcal{D}_i^F\}_{i=1, i \neq k}^K)\). We denote the global objective function as: \(\mathcal{L}_G(\theta) = \sum_{k=1}^K p_k \mathcal{L}_k(\theta)\)

\textbf{Assumption 1 (Smoothness).} Each loss function \(\mathcal{L}_k(\theta)\) is Lipschitz-smooth.

\textbf{Assumption 2 (Bounded Scalar).} \(\mathcal{L}_k(\theta)\) is bounded below by \(\mathcal{L}_{inf}\).

\textbf{Assumption 3 (Unbiased Gradient and Bounded Variance).} For each client, the stochastic gradient is unbiased: \(\mathbb{E}_\xi[g_k(\theta|\xi)] = \nabla \mathcal{L}_k(\theta)\), and has bounded variance \(\mathbb{E}_\xi[||g_k(\theta|\xi)-\nabla \mathcal{L}_k(\theta)||^2] \leq \sigma^2\).

\textbf{Assumption 4 (Bounded Dissimilarity).} For any set of weights $\{ p_k \geq 0 \}_{k=1}^K$ subject to $\sum_{k=1}^Kp_k=1$, there exists constants $\beta^2 \geq 1$ and $\kappa^2 \geq 0$ such that $\sum_{k=1}^K p_k||\nabla \mathcal{L}_k(\theta)||^2 \leq \beta^2 ||\nabla \mathcal{L}_G(\theta)||^2 + \kappa^2$.

Based on the above assumptions and the mathematical proof in \cite{ye2023fake}, we can prove the following theorem.

\textbf{Theorem 1 (Optimization bound of the global objective function).} Under these assumptions, if we set $\eta L \leq min\{ \frac{1}{2 E_{min}}, \frac{1}{\sqrt{2 E_{min} (E_{min}-1)(2 \beta^2 +1)}} \}$, 
The optimization error is bounded as follows:
\begin{equation}
\begin{aligned}
& \mathop{\min}_{t} \mathbb{E} \left\| \nabla \mathcal{L}_G(\theta^{(t)}) \right\|^2 \leq \frac{1}{T} \sum_{t=0}^{T-1} \mathbb{E} \left\| \nabla \mathcal{L}_G(\theta^{(t)}) \right\|^2 \\
& \leq \frac{4\left[ \mathcal{L}_G(\theta^{(T_{\beta}^{\text{turn}})}) - \mathcal{L}_{inf} \right]}{E_{min}\eta T} + 4 \eta L \sigma^2 \sum_{k=1}^K p_k^2 + \\ & 3(E_{min}-1)\eta^2\sigma^2L^2  
+ 6E_{min}(E_{min}-1)\eta^2L^2\kappa^2,
\end{aligned}
\end{equation}
where \(\eta\) is learning rate for local model training and \(E_{min}\) is the number of local model updates when communication round \(t > T_{\beta}^{\text{turn}}\).

\section{Experiments}
Our experiments is divided into five parts: a) the main setup of the experiments, b) the overall performance of FedEDS on different datasets and under varying levels of data heterogeneity, c) the communication efficiency of FedEDS, d) the training cost of FedEDS, and e) ablation studies of FedEDS.

\subsection{Experimental Setup}
\subsubsection{Federated Non-IID Datasets}
\begin{figure*}[t]
\centering
\includegraphics[width=1.0\textwidth]{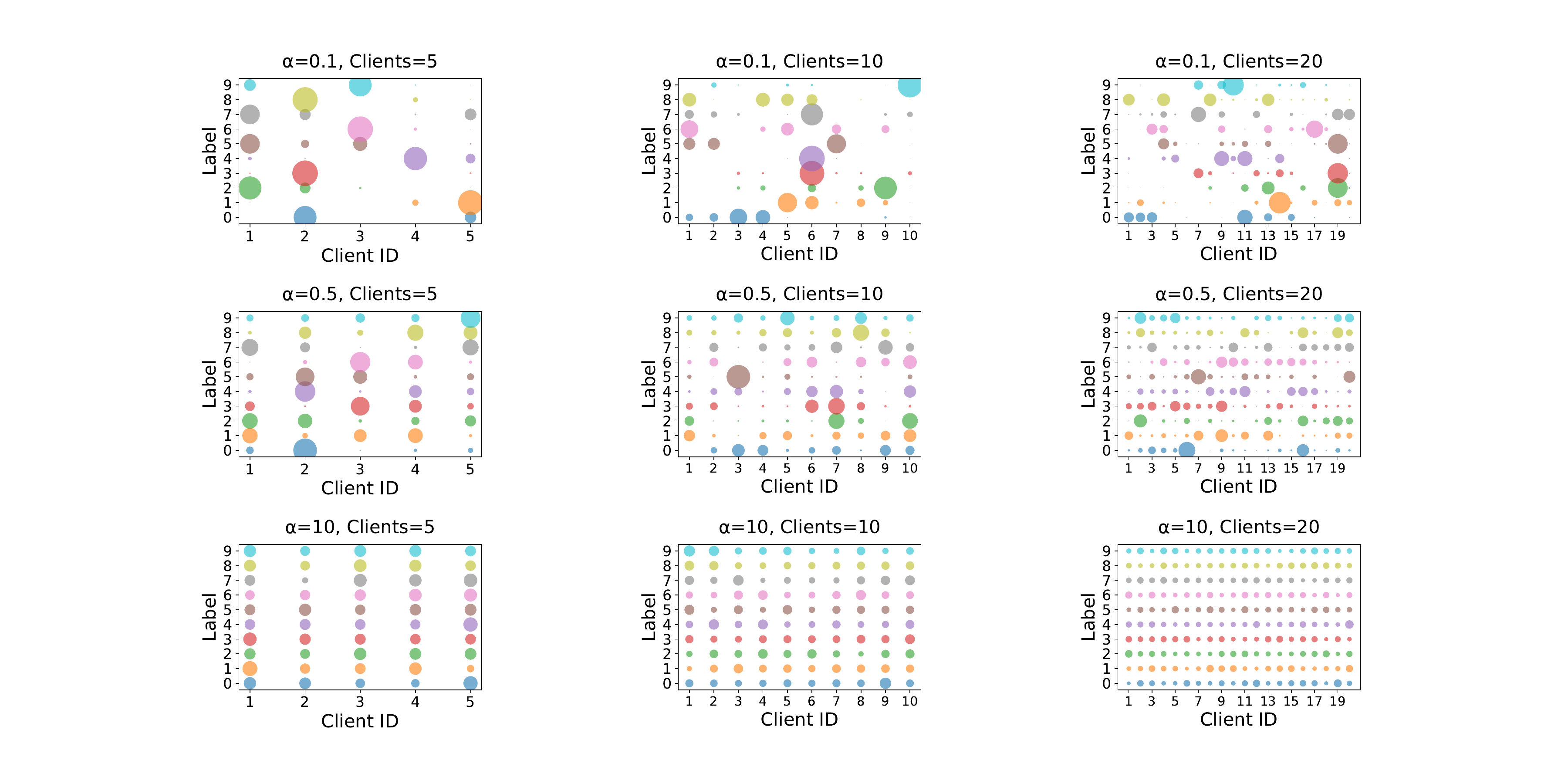} 
\caption{The visualizations of the data partitioning utilized in the experiments are presented, where the \(x\)-axis represents client IDs, the \(y\)-axis denotes the class labels, and the size of the scattered points corresponds to the number of training samples associated with each label available to the respective client.}
\label{fig5}
\end{figure*}

In this experiment, we use the Dirichlet distribution\cite{hsu2019measuring} to simulate the data heterogeneity of clients in real-world scenarios. We apply the Dirichlet distribution to partition the CIFAR-10 \cite{krizhevsky2009learning} and FashionMNIST\cite{xiao2017fashion} datasets with varying levels of heterogeneity. The heterogeneity levels are set as \( \alpha = \{0.1, 0.5, 10\} \), where \( \alpha \) corresponds to high, medium, and low levels of data heterogeneity, respectively. 
The first, second, and third rows of Fig.~\ref{fig5} illustrate the dataset distributions under high, medium, and low data heterogeneity, respectively, for different numbers of clients. As the level of data heterogeneity in federated learning decreases, the dataset distribution gradually becomes more uniform.

\subsubsection{Models, Metrics and Baselines}
The client model is based on the classic ResNet18 architecture \cite{he2016deep}, but with significant modifications to better suit the experimental setup. Specifically, all batch normalization layers are removed to eliminate their impact on the experiment. Additionally, a stochastic layer is incorporated within the network, with a configurable parameter that determines whether it participates in the forward pass. Furthermore, standard data augmentation techniques are disabled on the client side to ensure a controlled evaluation environment. The data encryptor model is a UNet\cite{ronneberger2015u}. The UNet\cite{ronneberger2015u} architecture consists of an encoder-decoder structure with skip connections to preserve spatial information. The encoder comprises two downsampling blocks, each consisting of a max-pooling layer followed by a double convolutional module. The double convolutional module applies two consecutive 3×3 convolutions with batch normalization and ReLU activation\cite{nair2010rectified}, progressively extracting hierarchical features. The decoder utilizes two upsampling blocks, where transposed convolutional layers restore spatial resolution, and concatenation with corresponding encoder features facilitates precise reconstruction. The final output is generated through a 1×1 convolutional layer, ensuring channel transformation to match the target output. 

We evaluate the model performance using two widely adopted metrics in Federated Learning: (a) communication rounds and (b) top-1 accuracy. As a plug-in approach, we apply FedEDS to prevailing existing FL algorithms, such as FedAvg\cite{mcmahan2017communication}, FedProx\cite{li2020federated}, and FedNova\cite{wang2020tackling} to compare the efficiency of our method.

\subsubsection{Experimental Details}
The experiments are conducted on an Nvidia RTX 3090 GPU. We set \(E_{max} = 5\), \(E_{min} = 1\), \(T_{\alpha}^{turn} = 1\), \(T_{\beta}^{turn} = 3\), \(m = 3\), \( \epsilon = 0.01\). The hyperparameter \(\mu\) of FedProx\cite{li2020federated} is set to 0.1. For the FashionMNIST\cite{xiao2017fashion} dataset, we set the total number of communication rounds to \( T = 30 \). For the CIFAR-10\cite{krizhevsky2009learning} dataset, \( T \) is set to 100 when \( K \in \{5, 10\} \) and 400 when \( K = 20 \). The client uses the SGD optimizer to train the local ResNet model and the AdamW optimizer to train the data encryptor model. The SGD optimizer is configured with a learning rate of 0.01, momentum of 0.9 to smooth updates and accelerate convergence, and weight decay of 0.0001 for L2 regularization to prevent overfitting. The AdamW optimizer is configured with a learning rate of 0.001.

\subsection{Performance Across Datasets and Data Heterogeneity}

Tables \ref{table1} and \ref{table2} present the accuracy performance of our proposed FedEDS method when integrated with FedAvg\cite{mcmahan2017communication}, FedProx \cite{li2020federated}, and FedNova\cite{wang2020tackling} on the CIFAR-10 \cite{krizhevsky2009learning} and FashionMNIST\cite{xiao2017fashion} datasets, respectively. The results demonstrate that FedEDS consistently improves model performance across varying numbers of clients \(K=\{5, 10, 20\}\) and different levels of data heterogeneity \(\alpha = \{0.1, 0.5, 10\}\).

For the CIFAR-10\cite{krizhevsky2009learning} dataset, FedEDS enhances the accuracy of all three baseline methods, particularly in highly heterogeneous settings (\(\alpha = 0.1\)), where the improvements are most pronounced. 
Specifically, when combined with FedAvg\cite{mcmahan2017communication}, FedProx\cite{li2020federated}, and FedNova\cite{wang2020tackling}, FedEDS achieves notable accuracy gains, with the largest improvement observed when \(K=20\) compared to their standalone counterparts. The trend holds across all client settings, indicating that FedEDS effectively mitigates the challenges posed by non-i.i.d. data distributions.

Similarly, for the FashionMNIST\cite{xiao2017fashion} dataset, FedEDS leads to consistent performance improvements. The most significant gains occur under moderate and high heterogeneity conditions, with improvements of up to 3.65\% when integrated with FedProx\cite{li2020federated} for \(K=20, \alpha=0.5\). The results suggest that FedEDS not only enhances convergence stability but also boosts generalization performance, particularly in federated learning environments with limited data diversity.

These findings underscore the robustness and efficacy of FedEDS in federated learning settings, demonstrating its adaptability across different datasets, client configurations, and levels of data heterogeneity. The consistent performance gains highlight its potential as a valuable enhancement to existing federated optimization strategies.

\begin{table*}[htbp]
\centering
\caption{The accuracy performance of using FedEDS combined with FedAvg\cite{mcmahan2017communication}, FedProx\cite{li2020federated}, and FedNova\cite{wang2020tackling}on the CIFAR-10\cite{krizhevsky2009learning} dataset under different levels of data heterogeneity and varying numbers of clients}
\label{table1}
\begin{tabular}{l|ccc|ccc|ccc}
\hline
Dataset & \multicolumn{9}{c}{CIFAR10} \\
\hline
Clients & \multicolumn{3}{c|}{\(K = 5\)} & \multicolumn{3}{c|}{\(K = 10\)} & \multicolumn{3}{c}{\(K = 20\)} \\
\hline
Method & \( \alpha = 0.1\) & \( \alpha = 0.5\) & \( \alpha = 10\) & \( \alpha = 0.1\) & \( \alpha = 0.5\) & \( \alpha = 10\) & \( \alpha = 0.1\) & \( \alpha = 0.5\) & \( \alpha = 10\)\\
\hline
FedAvg\cite{mcmahan2017communication} &69.24&73.04&75.11&68.91&71.63&73.44&68.87&68.91&71.64\\
FedProx\cite{li2020federated} &69.15&73.33&75.20&69.18&71.53&73.94&67.90&68.25&72.22\\
FedNova\cite{wang2020tackling}&69.24&73.16&75.61&68.85&71.89&73.91&67.59&68.63&72.73 \\
\hline
FedEDS + FedAvg\cite{mcmahan2017communication} &69.48 $\uparrow$&73.16 $\uparrow$&75.34 $\uparrow$&70.44 $\uparrow$&72.62 $\uparrow$&74.54 $\uparrow$&71.78 $\uparrow$&71.79 $\uparrow$&74.03 $\uparrow$\\
FedEDS + FedProx\cite{li2020federated} &70.41 $\uparrow$&73.70 $\uparrow$&76.14 $\uparrow$&71.13 $\uparrow$&73.51 $\uparrow$&74.81 $\uparrow$&71.54 $\uparrow$&72.18 $\uparrow$&74.66 $\uparrow$\\
FedEDS + FedNova\cite{wang2020tackling}&70.83 $\uparrow$&73.58 $\uparrow$&76.22 $\uparrow$&70.67 $\uparrow$&73.29 $\uparrow$&75.13 $\uparrow$&71.43 $\uparrow$&72.32 $\uparrow$&74.08 $\uparrow$\\
\hline
\end{tabular}
\end{table*}

\begin{table*}[htbp]
\centering
\caption{The accuracy performance of using FedEDS combined with FedAvg\cite{mcmahan2017communication}, FedProx\cite{li2020federated}, and FedNova\cite{wang2020tackling}on the FashionMNIST\cite{xiao2017fashion} dataset under different levels of data heterogeneity and varying numbers of clients}
\label{table2}
\begin{tabular}{l|ccc|ccc|ccc}
\hline
Dataset & \multicolumn{9}{c}{FashionMINST} \\
\hline
Clients & \multicolumn{3}{c|}{\(K = 5\)} & \multicolumn{3}{c|}{\(K = 10\)}& \multicolumn{3}{c}{\(K = 20\)} \\
\hline
Method & \( \alpha = 0.1\) & \( \alpha = 0.5\) & \( \alpha = 10\) & \( \alpha = 0.1\) & \( \alpha = 0.5\) & \( \alpha = 10\) & \( \alpha = 0.1\) & \( \alpha = 0.5\)& \( \alpha = 10\)\\
\hline
FedAvg\cite{mcmahan2017communication} &85.34&89.44&90.72&83.99&87.92&88.69&82.46&82.85&84.77\\
FedProx\cite{li2020federated} &86.23&88.47&90.15&82.29&86.85&88.08&81.90&82.58&84.37\\
FedNova\cite{wang2020tackling}&85.35&89.34&90.58&84.90&87.22&88.93&81.62&83.02&85.01\\
\hline
FedEDS + FedAvg\cite{mcmahan2017communication} &86.40 $\uparrow$&89.91 $\uparrow$&90.76 $\uparrow$&84.63 $\uparrow$&88.28 $\uparrow$&89.23 $\uparrow$&83.55 $\uparrow$&84.89 $\uparrow$&86.72 $\uparrow$\\
FedEDS + FedProx\cite{li2020federated} &86.98 $\uparrow$&90.04 $\uparrow$&90.74 $\uparrow$&84.19 $\uparrow$&88.27 $\uparrow$&89.58 $\uparrow$&83.99 $\uparrow$&86.23 $\uparrow$&87.60 $\uparrow$\\
FedEDS + FedNova\cite{wang2020tackling}&85.97 $\uparrow$&90.17 $\uparrow$&90.77 $\uparrow$&85.16 $\uparrow$&87.55 $\uparrow$&89.69 $\uparrow$&82.58 $\uparrow$&85.58 $\uparrow$&87.40 $\uparrow$\\
\hline
\end{tabular}
\end{table*}

\subsection{Communication Efficiency}
FedEDS can significantly reduce the number of communication rounds required for federated learning to converge. In our experiments, we compared the communication efficiency under different client numbers \(K = \{5, 10, 20\}\) and varying levels of data heterogeneity \(\alpha = \{0.1, 0.5, 10\}\), as shown in Table \ref{table3}, Table \ref{table4}, and Table \ref{table5}. The \ding{55} symbol in the table denotes that the target accuracy was not achieved. The experimental results demonstrate that FedEDS consistently decreases the communication rounds needed for convergence, and its acceleration effect becomes even more pronounced as the number of clients increases.

Compared with the FedAvg\cite{mcmahan2017communication}, FedProx\cite{li2020federated}, and FedNova\cite{wang2020tackling} methods used in the experiment, FedEDS has an additional process of client-side encrypted data transmission. In the real edge computing federated learning scenario, between each communication, the central server responsible for aggregating model parameters usually has to wait for a period of time (usually ranging from 1 minute to 10 minutes, depending on the network communication capabilities of the client and the central server) before receiving a response from the client\cite{li2022learning}. FedEDS targets scenarios where the edge servers participating in federated learning training are physically close and communication is faster, so the mutual transmission of encrypted data by FedEDS does not bring significant additional communication delays to federated learning.

\subsection{Training Cost}
Compared with the FedAvg\cite{mcmahan2017communication}, FedProx\cite{li2020federated}, and FedNova\cite{wang2020tackling} methods used in the experiment, FedEDS has an additional process of training the data encryptor. Training the data encryptor takes a certain amount of time, but FedEDS only trains the data encryptor once in the initial stage of federated learning, which means that FedEDS only increases the training time of edge servers in the initialization stage. We can estimate the impact of the training cost of this data encryptor on the training time of the entire federated learning. In our experiment, we used 20 epochs to train the data encryptor model. Since the settings of various parameters in the piecewise epoch annealing strategy have a small impact on the total number of epochs of all communication rounds of the client in the entire federated learning, we temporarily ignore the impact of the piecewise epoch annealing strategy on the total number of epochs. Assuming that the client trains the model for 1 epoch in each round of communication, we can calculate the total number of epochs required to achieve the target accuracy with and without FedEDS based on Table \ref{table3},Table \ref{table4},Table \ref{table5}. After calculation, we found that when the number of clients \(K = \{5,10\}\), the total number of epochs used by FedAvg\cite{mcmahan2017communication}, FedProx\cite{li2020federated}, and FedNova\cite{wang2020tackling} combined with FedEDS to achieve the specified accuracy is basically the same as that without FedEDS, which indicates that FedEDS does not increase the overall training computing cost of federated learning when \(K = \{5,10\}\). When the number of clients is \(K = 20\), FedEDS significantly reduces the total number of epochs used to achieve the specified accuracy, which indicates that FedEDS can significantly reduce the training computing cost of federated learning when the number of clients is large, which provides more possibilities for deploying federated learning on edge devices where computing resources are precious and limited.

\begin{table}[htbp]
\centering
\caption{The comparison of the communication rounds required to achieve the target accuracy between FedEDS and various baseline methods, including FedAvg\cite{mcmahan2017communication}, FedProx\cite{li2020federated}, and FedNova\cite{wang2020tackling}, on the CIFAR-10\cite{krizhevsky2009learning} dataset is presented. The total number of communication rounds for federated learning is \( T = 100 \), with the number of clients set to \( K = 5 \).}
\label{table3}
\begin{tabular}{c|l|cccccc}
\hline
\(\alpha\) & Method & 66\% & 67\% & 68\% & 69\% & 70\% \\
\hline
\multirow{6}{*}{0.1} & FedAvg\cite{mcmahan2017communication} & 40&40&56&71&\ding{55} \\
 & FedProx\cite{li2020federated} & 41&47&61&89&\ding{55}\\
 & FedNova\cite{wang2020tackling}& 38&41&53&99&\ding{55} \\
 & FedEDS + FedAvg\cite{mcmahan2017communication} & 27&30&35&41&\ding{55}\\
 & FedEDS + FedProx\cite{li2020federated} & 22&23&31&32&63\\
 & FedEDS + FedNova\cite{wang2020tackling}& 22&25&26&35&77 \\
\hline
\multirow{6}{*}{0.5} & FedAvg\cite{mcmahan2017communication} & 27&30&30&32&34\\
 & FedProx\cite{li2020federated} & 25&28&30&33&35 \\
 & FedNova\cite{wang2020tackling}& 23&26&28&31&32 \\
 & FedEDS + FedAvg\cite{mcmahan2017communication} & 19&19&21&21&25\\
 & FedEDS + FedProx\cite{li2020federated} & 15&18&18&20&21\\
 & FedEDS + FedNova\cite{wang2020tackling}& 14&15&18&19&22\\
\hline
\multirow{6}{*}{10} & FedAvg\cite{mcmahan2017communication} & 22&22&24&25&26 \\
 & FedProx\cite{li2020federated} & 23&23&25&26&28 \\
 & FedNova\cite{wang2020tackling}& 22&23&23&25&26 \\
 & FedEDS + FedAvg\cite{mcmahan2017communication} & 11&13&13&15&16\\
 & FedEDS + FedProx\cite{li2020federated} & 10&11&12&13&14\\
 & FedEDS + FedNova\cite{wang2020tackling}& 10&10&11&11&13\\
\hline
\end{tabular}
\end{table}

\begin{table}[htbp]
\centering
\caption{The comparison of the communication rounds required to achieve the target accuracy between FedEDS and various baseline methods, including FedAvg\cite{mcmahan2017communication}, FedProx\cite{li2020federated}, and FedNova\cite{wang2020tackling}, on the CIFAR-10\cite{krizhevsky2009learning} dataset is provided. The total number of communication rounds for federated learning is \( T = 100 \), with the number of clients set to \( K = 10 \).}
\label{table4}
\begin{tabular}{c|l|ccccc}
\hline
 \(\alpha\)& Method & 66\% & 67\% & 68\% & 69\% & 70\% \\
\hline
\multirow{6}{*}{0.1} & FedAvg\cite{mcmahan2017communication} & 66&70&74&\ding{55}&\ding{55}\\
 & FedProx\cite{li2020federated} & 66&71&75&93&\ding{55}\\
 & FedNova\cite{wang2020tackling}& 68&73&80&\ding{55}&\ding{55} \\
 & FedEDS + FedAvg\cite{mcmahan2017communication} & 34&38&43&51&88\\
 & FedEDS + FedProx\cite{li2020federated} & 27&30&34&37&59\\
 & FedEDS + FedNova\cite{wang2020tackling}& 27&31&35&38&47 \\
\hline
\multirow{6}{*}{0.5} & FedAvg\cite{mcmahan2017communication} & 53&56&58&61&66 \\
 & FedProx\cite{li2020federated} & 60&63&64&68&72 \\
 & FedNova\cite{wang2020tackling}& 55&57&61&65&70 \\
 & FedEDS + FedAvg\cite{mcmahan2017communication} & 24&24&27&31&33\\
 & FedEDS + FedProx\cite{li2020federated} & 22&23&24&27&30\\
 & FedEDS + FedNova\cite{wang2020tackling}& 20&22&24&27&30\\
\hline
\multirow{6}{*}{10} & FedAvg\cite{mcmahan2017communication} & 69&71&73&76&79 \\
 & FedProx\cite{li2020federated} & 68&70&72&74&78 \\
 & FedNova\cite{wang2020tackling}& 67&70&71&75&78 \\
 & FedEDS + FedAvg\cite{mcmahan2017communication} & 21&22&23&26&29\\
 & FedEDS + FedProx\cite{li2020federated} & 20&22&23&25&26\\
 & FedEDS + FedNova\cite{wang2020tackling}& 20&21&22&24&26\\
\hline
\end{tabular}
\end{table}

\begin{table}[htbp]
\centering
\caption{The comparison of the communication rounds required to achieve the target accuracy between FedEDS and various baseline methods, including FedAvg\cite{mcmahan2017communication}, FedProx\cite{li2020federated}, and FedNova\cite{wang2020tackling}, on the CIFAR-10\cite{krizhevsky2009learning} dataset is provided. The total number of communication rounds for federated learning is \( T = 400 \), with the number of clients set to \( K = 20 \).}
\label{table5}
\begin{tabular}{c|l|ccccc}
\hline
\(\alpha\) & Method & 66\% & 67\% & 68\% & 69\% & 70\% \\
\hline
\multirow{6}{*}{0.1} & FedAvg\cite{mcmahan2017communication} & 186&210&288&\ding{55}&\ding{55} \\
 & FedProx\cite{li2020federated} & 203&226&\ding{55}&\ding{55}&\ding{55}\\
 & FedNova\cite{wang2020tackling}& 224&329&\ding{55}&\ding{55}&\ding{55} \\
 & FedEDS + FedAvg\cite{mcmahan2017communication} & 60&65&73&83&115\\
 & FedEDS + FedProx\cite{li2020federated} & 58&64&70&88&131\\
 & FedEDS + FedNova\cite{wang2020tackling}& 52&58&65&78&97 \\
\hline
\multirow{6}{*}{0.5} & FedAvg\cite{mcmahan2017communication} & 270&274&296&\ding{55}&\ding{55} \\
 & FedProx\cite{li2020federated} & 223&241&311&\ding{55}&\ding{55} \\
 & FedNova\cite{wang2020tackling}& 240&247&269&\ding{55}&\ding{55} \\
 & FedEDS + FedAvg\cite{mcmahan2017communication} & 54&59&63&72&78\\
 & FedEDS + FedProx\cite{li2020federated} & 52&56&60&66&76\\
 & FedEDS + FedNova\cite{wang2020tackling}& 51&55&59&64&68\\
\hline
\multirow{6}{*}{10} & FedAvg\cite{mcmahan2017communication} & 288&296&303&309&321 \\
 & FedProx\cite{li2020federated} & 263&270&276&283&286 \\
 & FedNova\cite{wang2020tackling}& 257&263&267&273&279 \\
 & FedEDS + FedAvg\cite{mcmahan2017communication} & 43&45&48&53&55\\
 & FedEDS + FedProx\cite{li2020federated} & 40&42&45&49&52\\
 & FedEDS + FedNova\cite{wang2020tackling} & 41&44&46&50&53\\
\hline
\end{tabular}
\end{table}

\subsection{Ablation Study}
We conducted ablation studies on the proposed FedEDS method, using FedAvg\cite{mcmahan2017communication} as the baseline. The dataset used is CIFAR-10\cite{krizhevsky2009learning}, and under a data heterogeneity level of \( \alpha = 0.1 \), we tested with different numbers of clients to verify the contributions of two strategies: piecewise epoch annealing and encrypted data knowledge transfer. 

In Tables \ref{table6} and \ref{table7}, FedAvg\cite{mcmahan2017communication} + \(S_1\) represents the use of the piecewise epoch annealing strategy, and \(S_2\) represents the use of encrypted data knowledge transfer. Table \ref{table6} shows the contributions of strategies \(S_1\) and \(S_2\) to the Top-1 accuracy of federated learning under \(K = \{5, 10, 20\}\). Table \ref{table7} demonstrates the impact of \(S_1\) and \(S_2\) on the number of communication rounds required to achieve the target accuracy (i.e., convergence speed) under \(K = 20\).

\begin{table}[htbp]
    \centering
    \caption{The Impact of Knowledge Transfer by Encrypted Data Sharing and the Piecewise Epoch Annealing Strategy on the Top-1 Accuracy of Federated Learning.}
    \label{table6}
    \begin{tabular}{lccc}
        \hline
        Strategy & \( K = 5 \) & \( K = 10 \) & \( K = 20 \) \\
        \hline
        FedAvg\cite{mcmahan2017communication}       & 69.24 & 68.91 & 68.87 \\
        FedAvg\cite{mcmahan2017communication}+\(S_1\)    & 69.89 & 69.32 & 69.15 \\
        FedAvg\cite{mcmahan2017communication}+\(S_2\)    & 70.78 & 70.16 & 71.32 \\
        FedAvg\cite{mcmahan2017communication}+\(S_1\)+\(S_2\) & 70.83 & 70.67 & 71.43 \\
        \hline
    \end{tabular}
\end{table}

\begin{table}[htbp]
    \centering
    \caption{The Impact of Knowledge Transfer by Encrypted Data Sharing and the Piecewise Epoch Annealing Strategy on the Convergence Speed of Federated Learning.}
    \label{table7}
    \begin{tabular}{lccc}
        \hline
        Strategy & 50\% & 55\% & 60\% \\
        \hline
        FedAvg\cite{mcmahan2017communication}       & 130 & 145 & 175 \\
        FedAvg\cite{mcmahan2017communication}+\(S_1\)    & 99 & 114 & 143 \\
        FedAvg\cite{mcmahan2017communication}+\(S_2\)    & 29 & 41 & 56 \\
        FedAvg\cite{mcmahan2017communication}+\(S_1\)+\(S_2\) & 28 & 40 & 54 \\
        \hline
    \end{tabular}
\end{table}

Tables \ref{table6} and \ref{table7} illustrate that both the piecewise epoch annealing strategy (\(S_1\)) and encrypted data knowledge transfer (\(S_2\)), when applied individually, improve Top-1 accuracy and accelerate the convergence of federated learning compared to the baseline method. Moreover, the results indicate that encrypted data knowledge transfer (\(S_2\)) has a substantially greater impact on performance than the piecewise epoch annealing strategy (\(S_1\)). Notably, the combined application of \(S_1\) and \(S_2\) yields the most significant improvements, outperforming each approach when used in isolation.

\section{Conclusion}
In this work, we propose a novel framework named FedEDS. The core idea is to alleviate data distribution discrepancies (i.e., data heterogeneity) among clients in federated learning by sharing encrypted information.
FedEDS takes full advantage of high-speed transmission links between geographically proximate edge servers to share encrypted data, effectively addressing performance degradation caused by data heterogeneity in federated learning and accelerating its convergence. Furthermore, FedEDS can be easily integrated with other federated learning techniques to achieve better performance. FedEDS offers a new pathway to enhance federated learning training performance in edge scenarios while preserving privacy.

\section{Acknowledgements}
This work is supported by the National Natural Science Foundation of China grant No. 62032016 and 62372323, and the Foundation of Yunnan Key Laboratory of Service Computing under Grant YNSC24105. 

\bibliographystyle{IEEEtran}
\bibliography{IEEEexample}

\begin{thebibliography}{10}
\providecommand{\url}[1]{#1}
\csname url@samestyle\endcsname
\providecommand{\newblock}{\relax}
\providecommand{\bibinfo}[2]{#2}
\providecommand{\BIBentrySTDinterwordspacing}{\spaceskip=0pt\relax}
\providecommand{\BIBentryALTinterwordstretchfactor}{4}
\providecommand{\BIBentryALTinterwordspacing}{\spaceskip=\fontdimen2\font plus
\BIBentryALTinterwordstretchfactor\fontdimen3\font minus \fontdimen4\font\relax}
\providecommand{\BIBforeignlanguage}[2]{{%
\expandafter\ifx\csname l@#1\endcsname\relax
\typeout{** WARNING: IEEEtran.bst: No hyphenation pattern has been}%
\typeout{** loaded for the language `#1'. Using the pattern for}%
\typeout{** the default language instead.}%
\else
\language=\csname l@#1\endcsname
\fi
#2}}
\providecommand{\BIBdecl}{\relax}
\BIBdecl

\bibitem{mcmahan2017communication}
B.~McMahan, E.~Moore, D.~Ramage, S.~Hampson, and B.~A. y~Arcas, ``Communication-efficient learning of deep networks from decentralized data,'' in \emph{Artificial intelligence and statistics}.\hskip 1em plus 0.5em minus 0.4em\relax PMLR, 2017, pp. 1273--1282.

\bibitem{yang2019federated}
Q.~Yang, Y.~Liu, T.~Chen, and Y.~Tong, ``Federated machine learning: Concept and applications,'' \emph{ACM Transactions on Intelligent Systems and Technology (TIST)}, vol.~10, no.~2, pp. 1--19, 2019.

\bibitem{yang2020federated}
Q.~Yang, L.~Fan, and H.~Yu, \emph{Federated learning: Privacy and incentive}.\hskip 1em plus 0.5em minus 0.4em\relax Springer Nature, 2020, vol. 12500.

\bibitem{yi2024fedpe}
L.~Yi, X.~Shi, N.~Wang, J.~Zhang, G.~Wang, and X.~Liu, ``Fedpe: Adaptive model pruning-expanding for federated learning on mobile devices,'' \emph{IEEE Transactions on Mobile Computing}, 2024.

\bibitem{jiang2022fedmp}
Z.~Jiang, Y.~Xu, H.~Xu, Z.~Wang, C.~Qiao, and Y.~Zhao, ``Fedmp: Federated learning through adaptive model pruning in heterogeneous edge computing,'' in \emph{2022 IEEE 38th International Conference on Data Engineering (ICDE)}.\hskip 1em plus 0.5em minus 0.4em\relax IEEE, 2022, pp. 767--779.

\bibitem{he2020group}
C.~He, M.~Annavaram, and S.~Avestimehr, ``Group knowledge transfer: Federated learning of large cnns at the edge,'' \emph{Advances in Neural Information Processing Systems}, vol.~33, pp. 14\,068--14\,080, 2020.

\bibitem{Li_2020}
\BIBentryALTinterwordspacing
T.~Li, A.~K. Sahu, A.~Talwalkar, and V.~Smith, ``Federated learning: Challenges, methods, and future directions,'' \emph{IEEE Signal Processing Magazine}, vol.~37, no.~3, p. 50–60, May 2020. [Online]. Available: \url{http://dx.doi.org/10.1109/MSP.2020.2975749}
\BIBentrySTDinterwordspacing

\bibitem{lim2020federated}
W.~Y.~B. Lim, N.~C. Luong, D.~T. Hoang, Y.~Jiao, Y.-C. Liang, Q.~Yang, D.~Niyato, and C.~Miao, ``Federated learning in mobile edge networks: A comprehensive survey,'' \emph{IEEE communications surveys \& tutorials}, vol.~22, no.~3, pp. 2031--2063, 2020.

\bibitem{kairouz2021advances}
P.~Kairouz, H.~B. McMahan, B.~Avent, A.~Bellet, M.~Bennis, A.~N. Bhagoji, K.~Bonawitz, Z.~Charles, G.~Cormode, R.~Cummings \emph{et~al.}, ``Advances and open problems in federated learning,'' \emph{Foundations and trends{\textregistered} in machine learning}, vol.~14, no. 1--2, pp. 1--210, 2021.

\bibitem{li2021survey}
Q.~Li, Z.~Wen, Z.~Wu, S.~Hu, N.~Wang, Y.~Li, X.~Liu, and B.~He, ``A survey on federated learning systems: Vision, hype and reality for data privacy and protection,'' \emph{IEEE Transactions on Knowledge and Data Engineering}, vol.~35, no.~4, pp. 3347--3366, 2021.

\bibitem{zhao2018federated}
Y.~Zhao, M.~Li, L.~Lai, N.~Suda, D.~Civin, and V.~Chandra, ``Federated learning with non-iid data,'' \emph{arXiv preprint arXiv:1806.00582}, 2018.

\bibitem{li2021model}
Q.~Li, B.~He, and D.~Song, ``Model-contrastive federated learning,'' in \emph{Proceedings of the IEEE/CVF conference on computer vision and pattern recognition}, 2021, pp. 10\,713--10\,722.

\bibitem{li2020federated}
T.~Li, A.~K. Sahu, M.~Zaheer, M.~Sanjabi, A.~Talwalkar, and V.~Smith, ``Federated optimization in heterogeneous networks,'' \emph{Proceedings of Machine learning and systems}, vol.~2, pp. 429--450, 2020.

\bibitem{wang2020tackling}
J.~Wang, Q.~Liu, H.~Liang, G.~Joshi, and H.~V. Poor, ``Tackling the objective inconsistency problem in heterogeneous federated optimization,'' \emph{Advances in neural information processing systems}, vol.~33, pp. 7611--7623, 2020.

\bibitem{hsu2019measuringeffectsnonidenticaldata}
\BIBentryALTinterwordspacing
T.-M.~H. Hsu, H.~Qi, and M.~Brown, ``Measuring the effects of non-identical data distribution for federated visual classification,'' 2019. [Online]. Available: \url{https://arxiv.org/abs/1909.06335}
\BIBentrySTDinterwordspacing

\bibitem{zhang2022dense}
J.~Zhang, C.~Chen, B.~Li, L.~Lyu, S.~Wu, S.~Ding, C.~Shen, and C.~Wu, ``Dense: Data-free one-shot federated learning,'' \emph{Advances in Neural Information Processing Systems}, vol.~35, pp. 21\,414--21\,428, 2022.

\bibitem{zhu2021data}
Z.~Zhu, J.~Hong, and J.~Zhou, ``Data-free knowledge distillation for heterogeneous federated learning,'' in \emph{International conference on machine learning}.\hskip 1em plus 0.5em minus 0.4em\relax PMLR, 2021, pp. 12\,878--12\,889.

\bibitem{hao2021towards}
W.~Hao, M.~El-Khamy, J.~Lee, J.~Zhang, K.~J. Liang, C.~Chen, and L.~C. Duke, ``Towards fair federated learning with zero-shot data augmentation,'' in \emph{Proceedings of the IEEE/CVF conference on computer vision and pattern recognition}, 2021, pp. 3310--3319.

\bibitem{zhang2022fine}
L.~Zhang, L.~Shen, L.~Ding, D.~Tao, and L.-Y. Duan, ``Fine-tuning global model via data-free knowledge distillation for non-iid federated learning,'' in \emph{Proceedings of the IEEE/CVF conference on computer vision and pattern recognition}, 2022, pp. 10\,174--10\,183.

\bibitem{so2022lightsecagg}
J.~So, C.~He, C.-S. Yang, S.~Li, Q.~Yu, R.~E~Ali, B.~Guler, and S.~Avestimehr, ``Lightsecagg: a lightweight and versatile design for secure aggregation in federated learning,'' \emph{Proceedings of Machine Learning and Systems}, vol.~4, pp. 694--720, 2022.

\bibitem{jiang2022modelpruningenablesefficient}
\BIBentryALTinterwordspacing
Y.~Jiang, S.~Wang, V.~Valls, B.~J. Ko, W.-H. Lee, K.~K. Leung, and L.~Tassiulas, ``Model pruning enables efficient federated learning on edge devices,'' 2022. [Online]. Available: \url{https://arxiv.org/abs/1909.12326}
\BIBentrySTDinterwordspacing

\bibitem{jia2024dapperfldomainadaptivefederated}
\BIBentryALTinterwordspacing
Y.~Jia, X.~Zhang, H.~Hu, K.-K.~R. Choo, L.~Qi, X.~Xu, A.~Beheshti, and W.~Dou, ``Dapperfl: Domain adaptive federated learning with model fusion pruning for edge devices,'' 2024. [Online]. Available: \url{https://arxiv.org/abs/2412.05823}
\BIBentrySTDinterwordspacing

\bibitem{9660377}
S.~M. Shah and V.~K.~N. Lau, ``Model compression for communication efficient federated learning,'' \emph{IEEE Transactions on Neural Networks and Learning Systems}, vol.~34, no.~9, pp. 5937--5951, 2023.

\bibitem{zhu2024efficientmodelcompressionhierarchical}
\BIBentryALTinterwordspacing
X.~Zhu, S.~Yu, J.~Wang, and Q.~Yang, ``Efficient model compression for hierarchical federated learning,'' 2024. [Online]. Available: \url{https://arxiv.org/abs/2405.17522}
\BIBentrySTDinterwordspacing

\bibitem{xia2024bayesianfederatedmodelcompression}
\BIBentryALTinterwordspacing
C.~Xia, D.~H.~K. Tsang, and V.~K.~N. Lau, ``Bayesian federated model compression for communication and computation efficiency,'' 2024. [Online]. Available: \url{https://arxiv.org/abs/2404.07532}
\BIBentrySTDinterwordspacing

\bibitem{krizhevsky2009learning}
A.~Krizhevsky, G.~Hinton \emph{et~al.}, ``Learning multiple layers of features from tiny images,'' 2009.

\bibitem{ye2023fake}
R.~Ye, Y.~Du, Z.~Ni, S.~Chen, and Y.~Wang, ``Fake it till make it: Federated learning with consensus-oriented generation,'' \emph{arXiv preprint arXiv:2312.05966}, 2023.

\bibitem{hsu2019measuring}
T.-M.~H. Hsu, H.~Qi, and M.~Brown, ``Measuring the effects of non-identical data distribution for federated visual classification,'' \emph{arXiv preprint arXiv:1909.06335}, 2019.

\bibitem{xiao2017fashion}
H.~Xiao, K.~Rasul, and R.~Vollgraf, ``Fashion-mnist: a novel image dataset for benchmarking machine learning algorithms,'' \emph{arXiv preprint arXiv:1708.07747}, 2017.

\bibitem{he2016deep}
K.~He, X.~Zhang, S.~Ren, and J.~Sun, ``Deep residual learning for image recognition,'' in \emph{Proceedings of the IEEE conference on computer vision and pattern recognition}, 2016, pp. 770--778.

\bibitem{ronneberger2015u}
O.~Ronneberger, P.~Fischer, and T.~Brox, ``U-net: Convolutional networks for biomedical image segmentation,'' in \emph{Medical image computing and computer-assisted intervention--MICCAI 2015: 18th international conference, Munich, Germany, October 5-9, 2015, proceedings, part III 18}.\hskip 1em plus 0.5em minus 0.4em\relax Springer, 2015, pp. 234--241.

\bibitem{nair2010rectified}
V.~Nair and G.~E. Hinton, ``Rectified linear units improve restricted boltzmann machines,'' in \emph{Proceedings of the 27th international conference on machine learning (ICML-10)}, 2010, pp. 807--814.

\bibitem{li2022learning}
H.~Li, X.~Sun, and Z.~Zheng, ``Learning to attack federated learning: A model-based reinforcement learning attack framework,'' \emph{Advances in Neural Information Processing Systems}, vol.~35, pp. 35\,007--35\,020, 2022.

\end{thebibliography}
\end{document}